**Deep Reinforcement Learning Based Framework for Mobile Energy Disseminator Dispatching to Charge On-the-Road Electric Vehicles**

**Jiaming Wang[+]**
Graduate Research Assistant, Center for Connected and Automated Transportation (CCAT), Elmore Family School of Electrical and Computer Engineering, Purdue University, West Lafayette, IN, 47907.
Email: wang4146@purdue.edu
ORCID #: 0009-0005-2785-9303

**Jiqian Dong[+]**
Graduate Research Assistant, Center for Connected and Automated Transportation (CCAT), and Lyles School of Civil Engineering, Purdue University, West Lafayette, IN, 47907.
Email: dong282@purdue.edu
ORCID #: 0000-0002-2924-5728

**Sikai Chen**
Assistant Professor, Department of Civil and Environmental Engineering
University of Wisconsin-Madison, Madison, WI, 53706
Email: sikai.chen@wisc.edu
ORCID #: 0000-0002-5931-5619
 (Corresponding author)

**Shreyas Sundaram**
Marie Gordon Professor, Elmore Family School of Electrical and Computer Engineering
Co-Director, Institute for Control, Optimization and Networks (ICON)
Purdue University, West Lafayette, IN, 47907
Email: sundara2@purdue.edu
ORCID #: 0000-0002-5390-2505

**Samuel Labi**
Professor, Lyles School of Civil Engineering
Director, Center for Connected and Automated Transportation (CCAT), and
Purdue University, West Lafayette, IN, 47907.
Email: labi@purdue.edu
ORCID #: 0000-0001-9830-2071

+: equal contribution

*Wang, Dong, Chen, Sundaram, Labi*

**ABSTRACT**

The exponential growth of electric vehicles (EVs) presents novel challenges in preserving battery health and in addressing the persistent problem of vehicle range anxiety. To address these concerns, wireless charging, particularly, Mobile Energy Disseminators (MEDs) have emerged as a promising solution. The MED is mounted behind a large vehicle and charges all participating EVs within a radius upstream of it. Unfortunately, during such V2V charging, the MED and EVs inadvertently form platoons, thereby occupying multiple lanes and impairing overall corridor travel efficiency. Secondly, constrained budgets for MED deployment necessitate the development of an effective dispatching strategy to determine optimal timing and locations for introducing the MEDs into traffic. To address this problem, this paper proposes a deep reinforcement learning (DRL) based methodology to develop a vehicle dispatching framework. In the first component of the framework, we develop a realistic reinforcement learning (RL) environment termed "ChargingEnv" which incorporates a reliable charging simulation system that accounts for common practical issues in wireless charging deployment, specifically, the charging panel misalignment. The second component, the Proximal-Policy Optimization (PPO) agent, is trained to control MED dispatching through continuous interactions with ChargingEnv. Numerical experiments were carried out to demonstrate the demonstrate the efficacy of the proposed MED deployment decision processor. The results of the experiment suggest that the proposed model can significantly enhance EV travel range while efficiently deploying a optimal number of MEDs. The proposed model is found to be not only practical in its applicability but also has promises of real-world effectiveness. The proposed model can help travelers to maximize EV range and help road agencies or private-sector vendors to manage the deployment of MEDs efficiently.

## INTRODUCTION

### Benefit of Vehicle Electrification

The global automotive landscape has experienced a profound paradigm shift over the past few decades as the world grapples with the pressing challenges posed by climate change and the depletion of fossil fuels. As traditional internal combustion engine (ICE) vehicles continue to contribute significantly to greenhouse gas emissions and environmental degradation, there arises an urgent need for sustainable and eco-friendly transportation solutions. In response to this imperative, vehicle electrification has emerged as a transformative technology, offering a promising pathway towards achieving a greener and more sustainable future. Compared to the ICE, the key advantages of vehicle electrification are as follows:

- Zero Tailpipe Emissions: Electric Vehicles (EVs) boast a zero-emission profile, producing no tailpipe pollutants, CO2, or nitrogen dioxide (NO2). Although it is essential to acknowledge that battery production can contribute to the carbon footprint, the overall manufacturing process remains largely environmentally considerate (*1*). Even from a cradle-to-grave perspective, EVs are associated with significantly lower levels of greenhouse gas emissions compared to ICEVs.
- Lower range cost: Most trips by Americans involve short distances (92% of trips are 35 miles or less). Therefore, using a full-performance electric vehicle with a shorter range, where suitable, can significantly lower transportation costs. Electric vehicles hold the promise of being more cost-effective and greater energy security (*2*).
- Uncomplicated Maintenance: The mechanical composition of EVs is significantly simple, with fewer components and moving parts in the engine and mechanical parts. EVs do not have a cooling circuit, gearshift, clutch, or noise reduction components. This streamlined design results in cost-effective, less complex, and faster maintenance (*3*).
- Reducing Battery Cost: The cost of battery electric vehicles in the U.S. market is expected to drop to approximately $72/kWh in 2030. This is likely to result in electric vehicle cost parity with conventional vehicles between 2026–2028 for longer-range electric vehicles (*4*).
- Enhanced Comfort: The passenger experience in Electric Vehicles (EVs) is notably higher compared to ICEVs due to EV's minimal engine noise and vibration that provide a more comfortable journey.
- Superior Intelligent Features: Compared to conventional gasoline vehicles, EVs often feature advanced intelligent systems. These include enhanced connectivity (*5*), user interfaces, and autonomous driving features (*5–11*), reflecting the forefront of vehicular technology (*12*). Generally, a highly concentrated EV traffic flow is much easier to control than traditional human driven vehicles (HVs), which can facilitate global centralized controllers for a number of vehicles to achieve network level optimal decisions (*13*). For example, Du et. al, proposed a 2 stage-method combining GAQ (Graph Attention Network- Deep Q Learning) and EBkSP (Entropy Based k Shortest Path) to build a large-scale rerouting framework to improve travel efficiency (*14*) in the highly dynamic urban enviornment. Ha, et al., proposed a multiagent reinforcement learning for collaborative learning, is a promising solution to mitigate bottleneck traffic congestion (*15*). Furthermore, the EVs equipped with autonomous driving features can generally improve the road capacity since the headways between EVs are comparatively much smaller than HVs (*16*).

### EV Challenges

There exist numerous benefits to vehicle electrification. Yet still, there remains challenges that need to be addressed. The first challenge is that the battery cost constitutes the largest fraction of all component costs of an EV (Baek et al., 2022). At the current time, the experience of EV mass production is still not fully realized, resulting in limited (albeit slowly increasing) knowledge on battery degradation and life expectancy. Therefore, EVs currently have a higher depreciation rate compared to gasoline vehicles.

Another significant challenge, range anxiety (*17*), continues to pose a significant obstacle for electric vehicles (EVs). The limited driving range of EVs translates into the need for frequent charging which is inconvenient. Unlike conventional gasoline-powered vehicles, EVs cannot be quickly refueled at a station. Therefore, the EV traveler needs to carry out careful trip planning to ensure that the vehicle possesses adequate range to reach their destination or the next charging station. Such uncertainty regarding the range and the possibility of running out of power often causes stress and anxiety to EV owners. To address this challenge effectively, a promising approach that has been examined by researchers is dynamic wireless charging (DWC) (*18, 19*) that can provide electricity to EVs while they are in motion.

**Types of Dynamic Wireless Charging**

Dynamic Wireless Charging (DWC) technology builds upon the principles of wireless power transfer (WPT), specifically utilizing the Resonant Inductive Power Transfer (RIPT) method. It involves two separate coils that interact through oscillating magnetic fields without requiring any direct electrical connection. By enabling electric vehicles (EVs) to charge while in motion, DWC significantly reduces the need for frequent stops for charging, thereby alleviating range anxiety. For implementation of this technology, two solutions have been proposed in the literature: dynamic grid-to-vehicle (G2V) charging and vehicle-to-vehicle (V2V) charging.

*G2V: Grid to Vehicle*

The G2V approach involving charging the EV directly from the power grid infrastructure while the EV is in motion. The primary coil, embedded in the pavement, connects to the primary power grid through a combination of rectification, power factor correction, and high-frequency inversion circuits (*19*). These circuits collectively generate high-frequency magnetic fields that interact with the secondary coil installed at the base of the EV. **Figure 1 (a)** illustrates the principle of G2V, where the EV receives direct charging while driving over the in-pavement wireless charging panel.

Despite its effectiveness in addressing EV range anxiety, the G2V approach requires a comprehensive and expensive infrastructure to function effectively. First, to ensure broad coverage to avoid trip breaking by travelers, several road sections in a locality will need to be equipped with inductive charging pads. This involves costly reconstruction of the entire pavement or retrofitting existing pavements for installation. Secondly, since inductive charging pads need to be constantly activated along the entire road (or a lane), there is significant energy wastage. Fourth, it involves high cost of pavement and electric component maintenance. Fourth, power-seeking EVs need to travel to the nearest location with a G2V section, in order to charge. These drawbacks have hindered the widespread deployment of the G2V approach on a large scale.

*V2V: Vehicle to Vehicle*

A more dynamic and efficient on-the-drive charging scheme, known as vehicle-to-vehicle (V2V) charging, has been proposed in the literature. This works through the deployment of mobile energy disseminators (MEDs) (*20*). MEDs system set up can involve trucks or buses equipped with batteries that serve as energy sources to power neighboring EVs upstream of the MED (**Figure 1 (b)**). Unlike the G2V charging schema where EVs must go to specific locations (that is, the charging guideways) to charge, MEDs are portable and can be dispatched to deliver energy to multiple EVs regardless of their locations. This capability significantly enhances the mobility of EVs on the road. Moreover, utilizing MEDs has limited dependency on infrastructure, as there is no need to retrofit current road pavements with charging panels, making it a much easier mobile charging process.

However, a crucial question for implementing V2V charging in the real world remains unanswered: When and where should MEDs be dispatched? This consideration requires a balance between the driving range of EVs and the additional costs of introducing MEDs into the road network. Additionally, during the V2V charging phase, the MED and neighboring EVs seeking to be charged, will inadvertently form platoons and will thereby occupy multiple lanes. This situation could lead to significant impairment of the road network efficiency. To address this problem, this paper develops a vehicle dispatching framework to dispatch MEDs so that their service quality can be maximized.

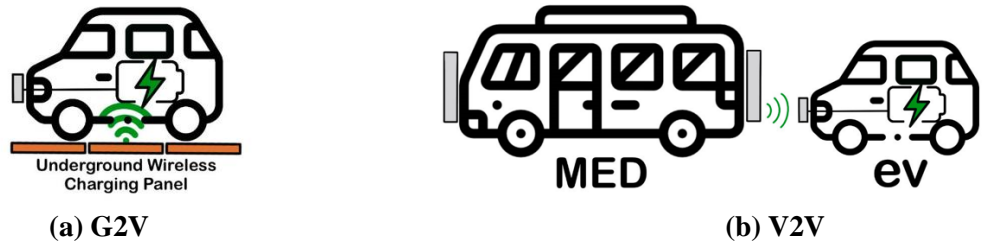

**(a) G2V**                    **(b) V2V**

**Figure 1 Different types of Dynamic Wireless Charging**

## RESEARCH GAPS AND MAIN CONTRIBUTIONS

There is a considerable body of research on vehicle dispatching with various goals, such as alleviating traffic congestion and enhancing logistics. However, there is limited past research on how to effectively dispatch mobile energy disseminators (MEDs) to improve the utility of electric vehicles (EVs) without compromising the utility of other road users, including gas vehicles or EVs that do not require charging at a specific time. To address this gap, our work adopts a data-driven approach, specifically reinforcement learning (RL), to develop an intelligent controller for MED dispatching in a multi-ramp highway scenario.

Another research gap lies in the absence of an existing simulation platform capable of accurately simulating V2V charging behaviors. Specifically, there is no simulator that incorporates high-fidelity physical V2V charging models, EV battery models, and background traffic agents. This limitation has hindered the development of RL-based frameworks to tackle charging-related problems. To bridge this gap, we introduce an RL environment called "ChargingEnv," built upon the Highway Env (*21*), to simulate the interactions among multiple MEDs, EVs, and background traffic agents on a highway scenario. Additionally, two realistic physical models are embedded in ChargingEnv, considering EV battery consumption and panel misalignment during V2V charging, to ensure high-fidelity simulation of EV battery and V2V charging dynamics. With its robustness, adaptability, and visualization capabilities, ChargingEnv serves as a versatile research platform to develop RL-based algorithms in mixed traffic scenarios involving EVs, MEDs, and other traffic agents.

In summary, the main contributions of this paper are as follows:

- Development of a robust and adaptable RL environment to simulate interactions between MEDs and EVs.

- Development of realistic physical models governing EV battery behavior and general V2V charging dynamics.

- Proposal of an RL-based framework to control the dispatching of MEDs, achieving a balance between extending EV range, minimizing dispatching costs, and minimizing traffic disruptions.

The rest of the paper is organized as follows: the Methods section introduces the details of the models used in the simulator "ChargingEnv" and the reinforcement learning basics for MED dispatching. The Experiment Setting section documents the implementation details of the proposed RL model and the interaction logic of vehicles within the environment. Then, the Results section compares the proposed model with the baselines and finally the paper concludes with the Conclusion.

## METHODS

### Transfer Efficiency in V2V Charging

In the highly dynamic driving scenario, during the V2V charging phase, the distances between the mobile energy disseminator (MED) and surrounding EVs constantly change. For example, when an EV is driven by a human and positioned to the left side of the MED, its speed may fluctuate, leading to inevitable horizontal positional misalignment between the transmitter and receiver coils. This misalignment has a significant impact on the efficiency of electricity transfer (*22, 23*). Essentially, an increase in lateral misalignment results in reduced mutual inductance, leading to a decrease in the transmitted power. To accurately quantify this effect and simulate the charging efficiency in our experiments, we have developed a realistic V2V charging model that takes into account the misalignment between the charging panels.

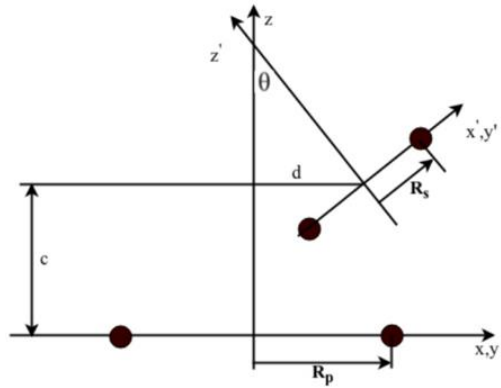

**Figure 2. Coil misalignment model, figure from (*22*)**

Before computing the charging efficiency of the V2V charging model, first we need to determine the mutual inductance, *M* using equations from the literature (*24, 25*), as follows:

$$M = \frac{\mu_o}{\pi}\sqrt{R_{MED}R_{EV}}\int_0^\pi \frac{\left[\cos(\theta) - \frac{d}{R_{EV}}\cos(\phi)\right]\psi(k)}{\sqrt{V^3}}d\phi \qquad (1)$$

$$V = \sqrt{1 - \cos^2(\phi)\sin^2(\theta) - 2\frac{d}{R_S}\cos(\phi)\cos(\theta) + \frac{d^2}{R_S^2}} \qquad (2)$$

where the parameters are:

$$k^2 = \frac{4\alpha V}{(1 + \alpha V)^2 + \xi^2}, \xi = \beta - \alpha\cos(\phi)\sin(\theta)$$

$$\alpha = \frac{R_{EV}}{R_{MED}}, \beta = \frac{c}{R_{MED}}, \qquad (3)$$

$$\Psi(k) = \left(\frac{2}{k} - k\right) K(k) - \frac{2}{k} E(k) = Q_{1/2}(x), x = \frac{2-k^2}{k^2} \tag{4}$$

$$M \propto N_{MED} N_{EV} \tag{5}$$

The parameter $\mu_o$ is the magnetic permeability of the coil inside charging panel, $R_{MED}$ represents MED coil radius of the coil inside charging panel. $R_{EV}$ is the EV coil radius of the coil inside charging panel. $\mu_o$ is magnetic permeability, $Q_{1/2}(x)$ represents Legendre function of the second kind and half-integral degree. The quantities $N_{MED}$ and $N_{EV}$ are the number of MED and EV coil turns, respectively, and $d, \theta, c$ are the horizontal misalignment, angular misalignment, and lateral misalignment of two coils.

In Equations 1-5, the charging coil is considered as the smallest decomposable unit to calculate mutual inductance. Of the various variables involved, the simulator can retrieve three types of misalignments during each step: Lateral Misalignment, Angular Misalignment, and Horizontal Misalignment. On the other hand, users have the flexibility to configure the remaining variables based on their specific requirements, such as the number of MED coil turns, magnetic permeability of the coil inside the charging panel, and more. Then, based on the mutual instance $M$, the model also takes the following, to output the charging efficiency $\eta$ in **Equation. 6**:

1. Load impedance ($Z_L$)
2. Parasite resistance of MED and EV (such as the built-in heater/AC inside the vehicle, $R_{MED}, R_{EV}$)
3. Resonant frequency ($\omega_0$)

All the three variables in this part are configured by the user.

$$\eta = \frac{Z_L}{(R_{MED}+Z_L)\left(1+\frac{R_{MED}(R_{EV}+Z_L)}{(\omega_0 M)^2}\right)} \tag{6}$$

**Physics Model for EV Battery Consumption**

To further simulate the relationship between vehicle speed and battery usage while moving at high speeds more accurately, an EV battery consumption model is developed. The model maps the mileage that the EV travels to the actual battery status. In other words, it governs "how much energy is consumed at each time step" when the EV travels at a certain speed, transfering the mileage achieved to the battery consumption model as a built-in model **Equation. 7-8** (*26*). This addition enhances the realism and validity of simulation environment, ensuring a closer approximation to real-world conditions.

For each timestep, the battery consumption in electric vehicles (EVs) is computed based on the tractive power, which is the power used by the vehicle to generate forward motion. A corresponding reduction in battery charge is then accounted for based on this calculated consumption.

$$P_{\text{Tractive}} = \sum_{i=1}^{i=N} \left[ mgC_{RR} + \frac{1}{2}\rho C_D A_F \left(\frac{V_i + V_{i-1}}{2}\right)^2 + mr\left(\frac{V_i - V_{i-1}}{t_{\text{inc}}}\right) + mg(\sin\theta) \right] * \left[\frac{V_i + V_{i-1}}{2}\right] \tag{7}$$

$$E = P_{\text{Tractive}} \times t_{\text{inc}}, \tag{8}$$

where $m$ is the mass of the vehicle, $g$ refers to gravitational acceleration, $C_{RR}$ refers to rolling resistance coefficient of the tyres used, $C_D$ refers to aerodynamic resistance thus depends on the vehicle's shape, $A_F$ refers to aerodynamic resistance thus depends on the vehicle's frontal area, $\rho$ refers to the density of surrounding air, $V_i$ refers to velocity at the end of the time span, $V_{i-1}$ refers to velocity at the start of the time span, $t_{inc}$ refers to time interval, $r$ is the rotate compensation, $\Theta$ refers to slope at the current road, and, E refers to battery consumption in the given time interval.

**Reinforcement Learning and Proximal-Policy Optimization (PPO)**

As shown in **Figure 3** (reinforcement learning), an agent could explore the environment and subsequently learn a behavior that promotes desired outcomes and avoids undesired outcomes (*27*). In this process, the agent observes the current states, takes action, and receives feedback (a positive or negative reward) from the environment (which is *ChargeEnv*, in the context of this paper). The agent evaluates the feedback signal, and understands the benefits (positive reward) of good actions and the (negative reward) of errant actions. In this paper, we use reinforcement learning to determine the dispatching logic of MEDs in a highway scenario, to help decide where and when to release a fully charged MED to charge EVs.

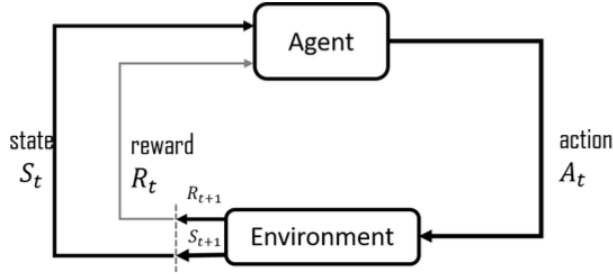

**Figure 3. Reinforcement Learning Concepts**

In this work, we adopt the Proximal-Policy Optimization (PPO) (*28*) as our RL agent. PPO is a well-known example of the policy gradient method which provides proof concerning monotonicity in the policy improvement over training iterations. In essense, PPO optimizes the Clipped Surrogate Objective as follows:

$$J^{\text{CLIP}}(\theta) = \hat{\mathbb{E}}_t \big[ min\big(r_t(\theta)\hat{A}_t, \text{clip}\,(r_t(\theta), 1 - \epsilon, 1 + \epsilon)\hat{A}_t\big)\big] \qquad (9)$$

where $\hat{A}_t$ is the estimated advantage of a specific time step $t$ by execute action $a_t$ and $r_t(\theta) = \pi_\theta(a_t \mid s_t)/\pi_{\theta_{old}}(a_t \mid s_t)$ denotes the probability ratio between the updated policy and old policy for generating the same action $a_t$. This surrogate objective function is adopted to prevent the policy from excessive variation in each iteration (when the absolute value of $\hat{A}_t$ is too large). Intuitively, this limits the update of policy function within the range of $[1 - \epsilon,\ 1 + \epsilon]$.

**EXPERIMENT SETTINGS**

**Scenario Settings**

*Road Network*

The proposed RL algorithm undergoes training and evaluation using example scenarios depicted in **Figure 4**. Specifically, we construct a 4-lane highway segment with 4 ramps, as illustrated in the same figure. The total length of the road is 4000 units, with each unit calibrated in HighwayEnv. The ramps are positioned at 100 units, 1200 units, 2100 units, and 3200 units, respectively. The unit measurement can be converted to any real-world calibration, such as miles or kilometers, to simulate real-world conditions accurately. To better reflect battery consumption conditions, we internally convert the units to meters, matching the battery consumption model within the environment.

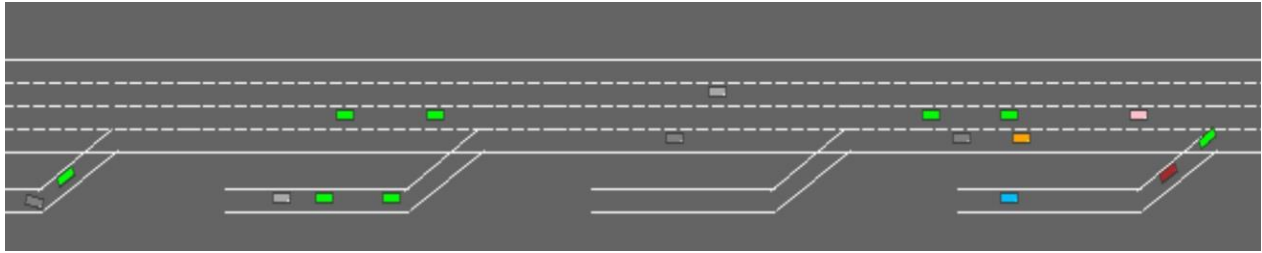

**Figure 4. Experimental Highway Scenario (Green: MEDs, Gray: background gas vehicles, other colors: individual EVs)**

*Vehicle Types and Control Logics*

To achieve a realistic simulation of the interaction between the MEDs and EVs, we introduce three types of vehicles into the environment: mobile energy disseminators (MEDs), electrical vehicles (EVs - both human-driven and fully autonomous), and normal gas vehicles (background traffic). At each simulation step, EVs and gas vehicles are randomly introduced from all four ramps and the left end of the highway, while the RL agent controls the timing and location of MEDs' release. All vehicles possess basic abilities, including overtaking, changing lanes, maintaining safe distances, and hazard avoidance, contributing to a more realistic scenario.

The vehicle control logics encompass both longitudinal and lateral control. For longitudinal control, we employ the Intelligent Driver Model (IDM) (*29*), ensuring collision-free driving for all vehicles. Lateral control governs the lane-changing behavior of vehicles. We primarily adopt the MOBIL model (*30*) for normal driving behaviors in this work. However, when EVs need to recharge, they must reach the designated charging slots provided by MEDs, making lane-changing mandatory during these specific scenarios.

*Charging and Battery Specifics*

The MED utilized in this study is equipped with four charging slots (charging panels) positioned on its left, right, rear, and front sides, allowing it to serve up to four EVs simultaneously. During charging, the MED and the surrounding charging EVs form a small platoon that occupies three lanes, as depicted in **Figure 5**. The table in **Figure 5** illustrates the service status of each slot on the MED, comprising two Boolean flags: "is booked" (True/False) and "is charging" (True/False).

When an EV's battery reaches a low state of charge (SoC), it sends a charging request to the nearest MED. If the MED has available charging slots and sufficient SoC to provide to the EV, it handles the request and returns the slot index to the EV, reserving the slot for charging. Once the EV reaches the designated location and attaches to the target slot, the "is charging" flag for that specific slot is set to True, initiating the charging process.

During the charging phase, EVs are expected to maintain a constant distance and speed with respect to the target MED. However, considering that EVs can be driven by humans, dynamic speed differences are natural to occur. This dynamic non-zero speed difference can lead to misalignment between the two charging panels, further reducing charging efficiency. To account for this, the V2V transfer efficiency model mentioned in the Methods section is incorporated. Once the charging is complete, the EVs resume their initial speed before charging and leave the charging area, creating space for subsequent EVs in need of charging.

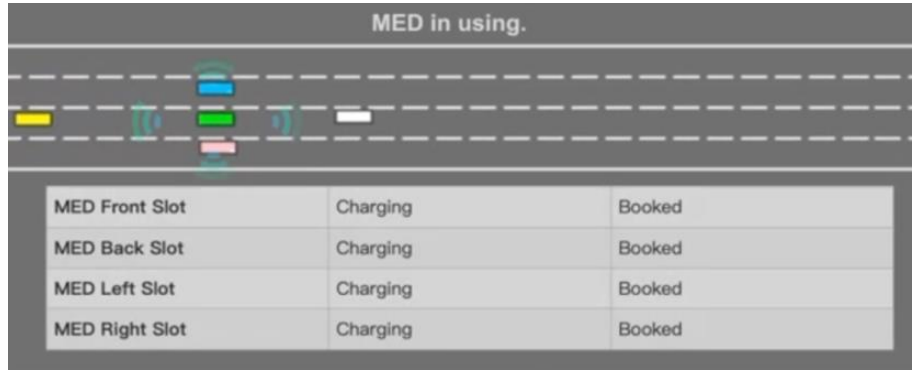

**Figure 5. Charging Arrangement**

**State Space**

As shown in **Figure 6**, the state space considered in this work consists of 2 parts, for MEDs and EVs respectively.

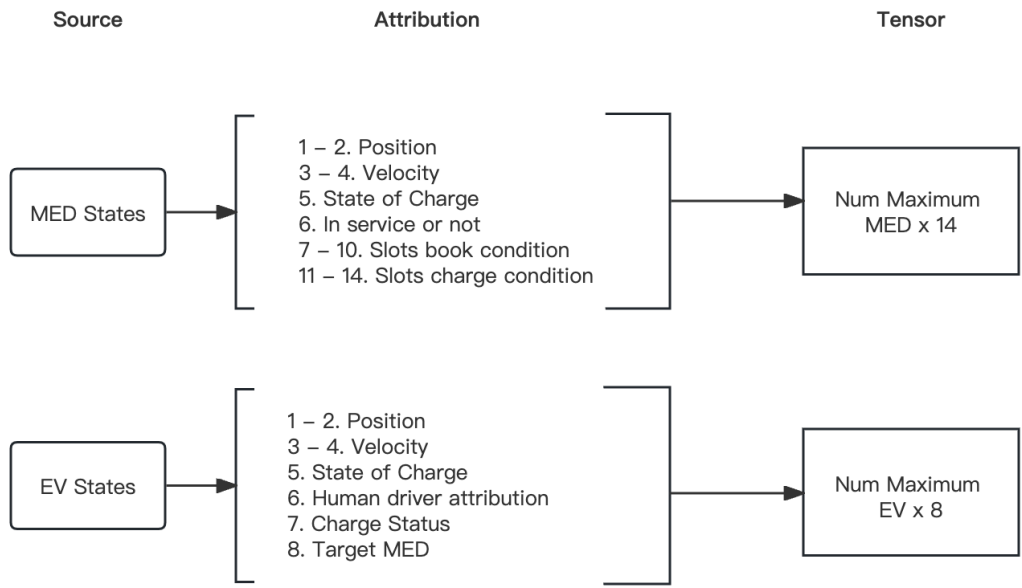

**Figure 6. State Space Representation**

Regarding the MEDs, the following six categories of information are considered:

1. The x-coordinate of the MED's position.

2. The y-coordinate of the MED's position.

3. The x-component of the MED's velocity.

4. The y-component of the MED's velocity.

5. The percentage of the battery remaining for the MED.

6. A boolean indicating if the MED is in service (1 if yes, 0 if no).

7. Booleans indicating if each of the four slots is booked (1 if yes, 0 if no).

8. Booleans indicating if each of the four slots is charging (1 if yes, 0 if no).

For all the EVs, similar information is included:

1. The x-coordinate of the EV's position.

2. The y-coordinate of the EV's position.

3. The x-component of the EV's velocity.

4. The y-component of the EV's velocity.

5. The percentage of the battery remaining for the EV.

6. A boolean indicating if the EV is driven by a human (1 if yes, 0 if no).

7. A boolean indicating the charging status of the EV (1 if charging, 0 if not).

8. The information of the target Medium Energy Device (MED) for the EV.

The overall state representation that is fed to the RL model is a flattened array containing all the information mentioned above.

**Action Space**

At each timestep, the agent can position the Mobile Electric Device (MED) on one of four ramps. The action space consists of discrete actions ranging from 0 to 4. Choosing 0 means the agent opts not to allocate any MED, while selecting values from 1 to 4 indicates the allocation of the MED to the respective four ramps.

Once placed, the MED will access the highway via the ramp and initiates charging of proximate vehicles once it moves onto the service lane. Additionally, there will be a user-defined interval known as the "cooldown period," during which the agent is restricted from releasing the vehicle. This cooldown period is intentionally included to prevent rapid and potentially detrimental fluctuations in resource allocation.

**Reward Settings**

In this paper, we consider three types of rewards, and one types of penalty: the destination reward, speed reward, lane-changing penalty, and collision penalty, as listed in **Table 1.**

**Table 1. Rewards and Penalties**

| Category | Description | Symbol |
|---|---|---|
| EV battery-depletion penalty | The "EV battery-depletion penalty" is a fixed number. If the battery level of an Electric Vehicle (EV) drops to zero while on the road, a significant amount of reward is deducted. This serves as a disincentive for the agent to allow the EVs to run out of battery, encouraging more effective management of resources and energy. | $P_{Depleted\_battery}$ |
| EV State of charge reward | The "EV average battery reward" is a variable that provides rewards based on the State of Charge (SoC) of EVs at the current time step. A higher SoC corresponds to a higher reward. This mechanism incentivizes the agent to maintain a higher average battery level among all EVs, encouraging more efficient energy management and prolonging the operation time of EVs. | $R_{SoC}$ |

| Distance Reward | The "distance reward" refers to the total distance covered by all electric vehicles (EVs) at the current time step. This reward system encourages the agent to maximize the total distance traveled by all EVs, which can be seen as promoting overall efficiency and utilization of the EV fleet. The more distance the EVs cover, the higher the reward, thus incentivizing optimal routing and energy management. | $R_d$ |
| --- | --- | --- |
| Speed reward | The "speed reward" is designed to prevent traffic congestion. It is typically calculated based on the average speed of all electric vehicles (EVs) in the system at a given time step. The higher the average speed (indicating less traffic congestion), the greater the reward. This encourages the agent to make decisions that lead to smoother traffic flow and less congestion, enhancing the overall efficiency of the EV fleet. | $R_v$ |

The overall reward function is defined as:

$$R_{total} = -w_1 P_{battery\_depletion} + w_2 R_{SoC} + w_3 R_d + w_4 R_v \qquad (10)$$

## RESULTS

### Training Curve

The model is trained for 400k steps, and the training curve (in terms of average episode reward) can be visualized in **Figure 7.** As shown in the curve, the model converges fast and consistently make progress during the training process.

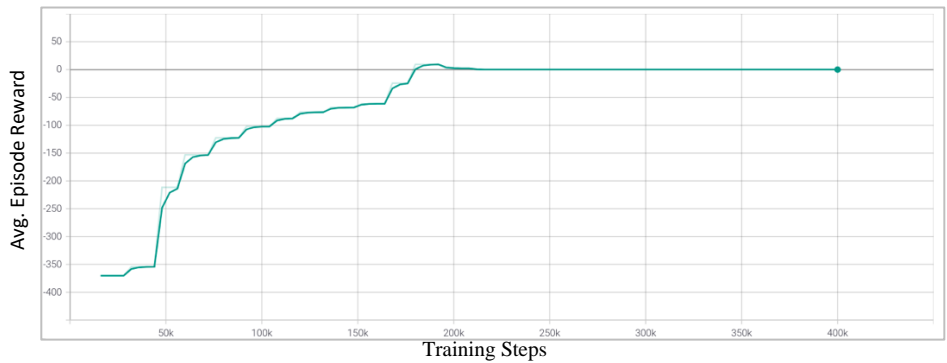

**Figure 7. The Training Curve**

### Comparative Results

In this study, we evaluate our proposed Deep Reinforcement Learning (DRL) approach for Multi-Electric Vehicle (EV) Deployment (MED). We compare our model against two baselines: random choice and no action. To assess the performance, we run the entire episode for 10 iterations and calculate the following statistics:

1. Mean of episode reward
2. Standard deviation of episode reward
3. Proportion of EVs experiencing battery depletion
4. Average EV traveling range (unit)
5. EV average state of charge (SoC)

The results for mean and standard deviation of episode rewards are presented in **Table 2**, while the other statistics are visualized in **Figure 8.** From these results, it can be easily observed that our proposed DRL model demonstrates remarkable improvements in EV performance compared to the baselines. Specifically, it reduces the chance of EV battery depletion by 50%, increases the average EV traveling range by 25%, and raises the EV state of charge (SoC) by 20%. These enhancements are achieved with a limited number of MEDs deployed (15 MEDs).

The outcomes show that our RL agent outperforms the baselines significantly. This success highlights the practicality of deploying our proposed model in real-world scenarios for MED deployment. The reason lies in the ability of the Proximal Policy Optimization (PPO) agent to learn a sophisticated metering logic based on traffic status, including EV density and their current SoCs. Consequently, the MEDs are efficiently dispatched to the areas where they are most needed. Such decision-making is challenging for human and rule-based approaches, given the high dimensionality and dynamic nature of the input data. The efficacy of our DRL-based approach in the MED dispatching task is evident, indicating the potential for its successful application in real-world scenarios.

**Table 2. Comparative results**

| Models | Avg. reward for each episode | Std. of reward for each episode |
|---|---|---|
| Random choice | -332.4056918 | 53.94543289 |
| No action | -521.874184 | 16.52914886 |
| PPO Agent | 0.340361448 | 0.141114801 |

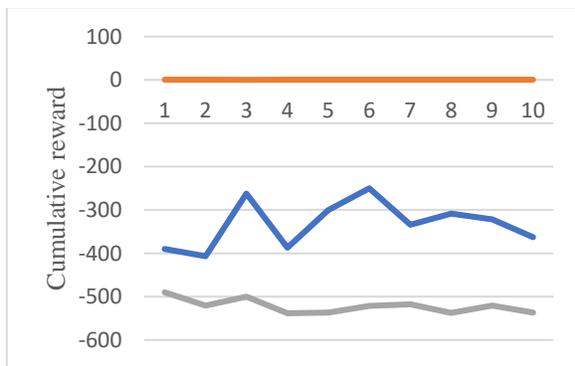

(a)Cumulative Reward

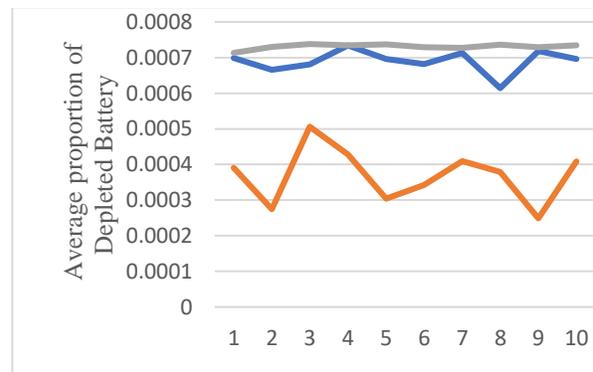

(b)Average Probability of Battery Depletion

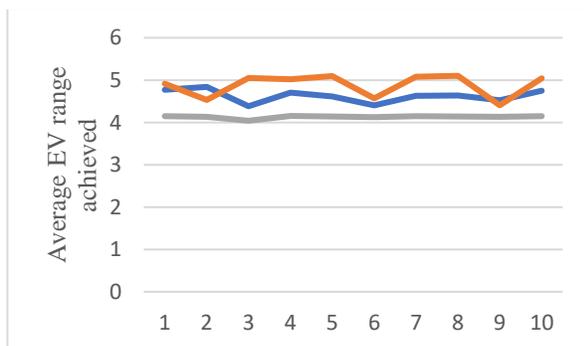

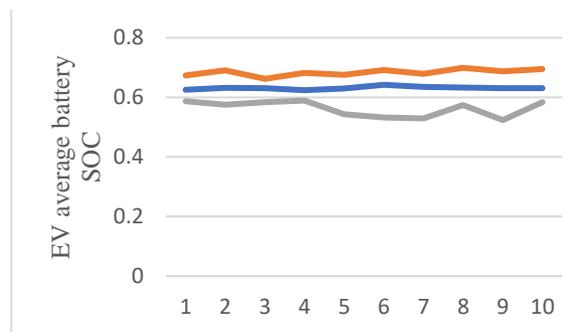

(c) Average EV Range Achieved    (d) EV Average Battery State-of-Charge

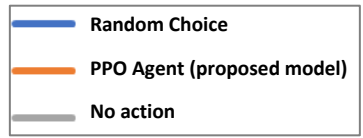

**Figure 8. Results of the experiment (comparison across the different models, 10-iterations)**

## CONCLUDING REMARKS

In conclusion, this paper addresses the challenges and opportunities in the realm of dynamic wireless charging (DWC) for electric vehicles (EVs). More specifically, we trained a PPO agent algorithm to serve as an intelligent controller for dispatching MEDs. The objective is to dispatch MEDs in a manner that achieves a balance between extending EV range, minimizing dispatching costs, and minimizing traffic disruptions. This is the first model that addresses the MED dispatching problem. In addition, we create a realistic V2V charging model that considers misalignment between charging panels, and thereby simulates real-world conditions more accurately. Our proposed reinforcement leaning environment, "ChargingEnv," incorporates high-fidelity physical models and serves as a versatile research platform for addressing the stated problems associated with on-the-go charging of EVs in a mixed traffic scenario.

   Moving forward, we foresee substantial improvements in our simulator, particularly regarding the action space of the agent and the implementation of multi-agent systems. In future iterations, we plan build a multi-agent control system to control the motion of every MED. This enhancement will equip the agents to adeptly handle requests emanating from Electric Vehicles. Compared to the task in the current experiment, this upgrade may significantly improve charging efficiency, thereby allowing us to explore the potential of MED applications on the highway in greater depth.

## ACKNOWLEDGMENTS

This work was supported by Purdue University's Center for Connected and Automated Transportation (CCAT), a part of the larger CCAT consortium, a USDOT Region 5 University Transportation Center funded by the U.S. Department of Transportation Office of the Assistant Secretary for Research and Technology (OST-R), University Transportation Centers Program Award #69A3551747105. Also, general support was provided by the Center for Innovation in Control, Optimization, and Networks (ICON) and the Autonomous and Connected Systems (ACS) initiatives at Purdue University's College of Engineering. The contents of this paper reflect the views of the authors, who are responsible for the facts and the accuracy of the data presented herein, and do not necessarily reflect the official views or policies of the sponsoring organization.

This manuscript is herein submitted for PRESENTATION ONLY at the 2024 Annual Meeting of the Transportation Research Board.

## DECLARATION OF CONFLICTING INTERESTS

The authors declare no potential conflicts of interest with respect to the research, authorship of this submission.

## AUTHOR CONTRIBUTIONS

The authors confirm contribution to the paper as follows: all authors contributed to all sections. All authors reviewed the results and approved the final version of the manuscript.